\documentclass[letter, 10 pt, conference]{ieeeconf}  

\IEEEoverridecommandlockouts 
\usepackage[english]{babel}
\usepackage[utf8]{inputenc}
\usepackage{times} 
\usepackage{cite}

\DeclareMathAlphabet{\pazocal}{OMS}{zplm}{m}{n}

\usepackage{hyperref}
\usepackage[nolist]{acronym}

\usepackage{multirow}

\usepackage{siunitx}
\usepackage{graphicx} 
\usepackage{epsfig} 
\usepackage{pgfplots}
\usepackage{caption}
\usepackage[font=footnotesize]{subcaption}
\usepackage{pgfplots}

\usepackage{float}

\usepackage[export]{adjustbox}

\usepackage{xcolor}
\definecolor{myYellow}{rgb}{0.93,0.69,0.13}
\definecolor{myPurple}{rgb}{0.49,0.18,0.56}
\definecolor{myGreen}{rgb}{0.26 0.72 0.54}

\usepackage{mathtools}
\usepackage{nicefrac}
\usepackage{amsmath}
\usepackage{amssymb}
\usepackage{bm} 

\DeclareMathOperator*{\argmax}{argmax}

\newcommand{\mset}[1]{\mathcal{#1}}

\usepackage{etoolbox}
\makeatletter%
\AfterPreamble{%
	\usepackage{hyperref}%
	\let\oldhypertarget\hypertarget%
	\renewcommand{\hypertarget}[2]{%
		\oldhypertarget{#1}{#2}%
		\protected@write\@mainaux{}{%
			\string\expandafter\string\gdef%
			\string\csname\string\detokenize{#1}\string\endcsname{#2}%
		}%
	}%
	\newcommand{\myhyperlink}[1]{%
		\hyperlink{#1}{\csname #1\endcsname}%
	}%
}
\makeatother%

\newcounter{Remark}
\newcounter{Definition}
\newcounter{Problem}
\usepackage{algorithm}
\usepackage[noend]{algpseudocode}

\makeatletter
\def\BState{\State\hskip-\ALG@thistlm}
\makeatother

\usepackage{tikz}
\pgfdeclarelayer{foreground}
\pgfsetlayers{background, main, foreground}
\usetikzlibrary{quotes, angles, backgrounds, arrows, automata, shapes, positioning, calc, through, spy, decorations.pathreplacing, decorations.markings, arrows.meta, automata, petri}

\tikzset{
    imglabel/.style={
      rectangle,
      inner sep=2pt,
      text=black,
      minimum height=1em,
      text centered,
      fill=white,
      fill opacity=1.0,
      text opacity=1,
      anchor=south west,
    },
  }
\tikzset{
	state/.style={
		rectangle,
		draw=black, very thick,
		minimum height=1.0em,
		text centered,
	},
}
\tikzset{
  on each segment/.style={
    decorate,
    decoration={
      show path construction,
      moveto code={},
      lineto code={
        \path [#1]
        (\tikzinputsegmentfirst) -- (\tikzinputsegmentlast);
      },
      curveto code={
        \path [#1] (\tikzinputsegmentfirst)
        .. controls
        (\tikzinputsegmentsupporta) and (\tikzinputsegmentsupportb)
        ..
        (\tikzinputsegmentlast);
      },
      closepath code={
        \path [#1]
        (\tikzinputsegmentfirst) -- (\tikzinputsegmentlast);
      },
    },
  },
  mid arrow/.style={postaction={decorate,decoration={
        markings,
        mark=at position .5 with {\arrow[#1]{stealth}}
      }}},
}

\graphicspath{{./figures/}}

\newcommand\copyrighttext{%
    \small \begin{center} \color{red} \textcopyright\,Accepted for presentation to the ``Breaking Swarm Stereotypes" Workshop at ICRA'24, Yokohama, Japan. Personal use of this material is permitted. Permission from authors must be obtained for all other uses, in any current or future media, including reprinting/republishing this material for advertising or promotional purposes, creating new collective works, for resale or redistribution to servers or lists, or reuse of any copyrighted component of this work in other works. \end{center}}
\newcommand\copyrightnotice{%
	\begin{tikzpicture}[remember picture,overlay]
	\node[anchor=south,yshift=25.6cm] at (current page.south) 
	{\color{red}\fbox{\parbox{\dimexpr\textwidth-\fboxsep-\fboxrule\relax}{\copyrighttext}}};
	\end{tikzpicture}%
}

\title{\copyrightnotice \LARGE \bf PACNav: Enhancing Collective Navigation for UAV Swarms in Communication-Challenged Environments}

\author{Afzal Ahmad$^{1}$, Daniel Bonilla Licea$^{1,2}$, Giuseppe Silano$^{1}$, Tomas Baca$^{1}$, and Martin Saska$^{1}$
    \thanks{$^1$Afzal Ahmad, Daniel Bonilla Licea, Giuseppe Silano, Tomas Baca, and Martin Saska are with the Department of Cybernetics, Czech Technical University in Prague, 12135 Prague, Czech Republic (email: {\tt name.surname@fel.cvut.cz}).}
    \thanks{$^2$Daniel Bonilla Licea is with the college of Computing, Mohammed VI Polytechnic University, Ben Guerir, Morocco (email: {\tt daniel.bonilla@um6p.ma}).}
    \thanks{$^\dag$This work was partially supported by the Czech Science Foundation (GAČR) grant no. 23-07517S, and by CTU grant no. SGS23/177/OHK3/3T/13.}
}

\begin{document}

\maketitle
\thispagestyle{empty} 
\pagestyle{empty} 


\begin{abstract}

This article presents~\ac{PACNav} as an approach for achieving decentralized collective navigation of~\ac{UAV} swarms. The technique is inspired by the flocking and collective navigation behavior observed in natural swarms, such as cattle herds, bird flocks, and even large groups of humans. \ac{PACNav} relies solely on local observations of relative positions of~\acp{UAV}, making it suitable for large swarms deprived of communication capabilities and external localization systems. We introduce the novel concepts of \textbf{path persistence} and \textbf{path similarity}, which allow each swarm member to analyze the motion of others.~\ac{PACNav} is grounded on two main principles: (1)~\acp{UAV} with little variation in motion direction exhibit high \textbf{path persistence} and are considered reliable leaders by other~\acp{UAV}; (2) groups of~\acp{UAV} that move in a similar direction demonstrate high \textbf{path similarity}, and such groups are assumed to contain a reliable leader. The proposed approach also incorporates a reactive collision avoidance mechanism to prevent collisions with swarm members and environmental obstacles. The method is validated through simulated and real-world experiments conducted in a natural forest.

\end{abstract}



\section{Full-version}
\label{sec:fullVersion}

A full version of this work is available at~\url{https://iopscience.iop.org/article/10.1088/1748-3190/ac98e6}. To reference, use~\cite{Afzal2022Biob}.



\section{Introduction}
\label{sec:introduction}

The use of a group of~\acp{UAV} can reduce mission time and provide the redundancy and safety that is critical in many real-world applications
~\cite{McGuireScienceRobotics2019}. However, employing a centralized system to control the motion of all the~\acp{UAV} in the swarm can be challenging due to the unavailability of reliable and real-time information about the environment and other~\acp{UAV} in the swarm. Animals, like fish and birds, serve as prime examples of multi-agent systems that employ decentralized decision-making for collective motion~\cite{vicsek_2010_nature, olfati_2006_tac}. For instance, \cite{olfati_2006_tac} draw insights from animal motion to devise a set of simple rules addressing attraction and repulsion to neighbors and alignment with the group, enabling collective motion. In many cases, decentralized decision-making systems rely solely on local information about neighbors, making these methods scalable to a large number of robots. 

This paper introduces a bioinspired decentralized approach for the collective navigation of a swarm of~\acp{UAV}, leveraging onboard sensor data for control without reliance on~\ac{GNSS} or communication. By demonstrating~\acp{UAV} navigating effectively through a forest environment, the approach challenges the stereotype of swarms being limited to ``toy scenarios,'' highlighting their potential for real-world applications. Drawing from collective motion analysis in animal and human groups, \textit{path similarity} and \textit{path persistence} metrics are designed to compare~\ac{UAV} trajectories. Individual~\acp{UAV} follow a target~\ac{UAV} selected based on these metrics, enabling collective motion. Emphasizing safety, the collision avoidance mechanism is tailored for complex real-world environments. Both simulated experiments and real-world flights in natural forest environments validate and analyze the approach's performance and robustness. The related source code has been released as open-source\footnote{\url{https://github.com/ctu-mrs/pacnav}\label{fotnote:code}}.



\section{Problem Description}
\label{sec:problem_description}

We address the challenge of navigating a~\ac{UAV} swarm, which lacks communication and global localization capabilities, within an environment containing randomly distributed obstacles. The objective of the swarm is to collectively advance towards a goal location known only to a subset of the~\acp{UAV}. Navigation relies on on-board sensors for mapping and localizing obstacles and other~\acp{UAV} within the environment.



\section{Problem Solution}
\label{sec:problem_solution}

We introduce a decentralized control method for~\acp{UAV} that rely solely on onboard sensors and computational resources to govern their motion. As the movement of each~\ac{UAV} is influenced by the motion of others in the swarm, collective navigation emerges through the control of individual~\acp{UAV}. Our method comprises of two phases: In the first phase, each~\ac{UAV} determines a suitable target~\ac{UAV} to follow, and in the second phase, it computes motion control commands to reach the target while avoiding collisions with obstacles and other~\acp{UAV}.

During the first phase, at each time instant $k$, the $i$-th~\ac{UAV} selects a target location $\mathbf{d}_i[k] \in \mathbb{R}^2$ and plans a path to reach it. This target can either be the goal position $\mathbf{g}$ if the~\ac{UAV} belongs to the informed subgroup or a neighboring~\ac{UAV} potentially moving towards the goal $\mathbf{g}$. To select this target, we generate a set of potential targets $\mathcal{T}_i[k]$, considering three criteria: (i) \textit{\acp{UAV} not in close proximity}:~\acp{UAV} close to the $i$-th~\ac{UAV} are mostly influenced by the collision avoidance mechanisms, so only those beyond a certain distance are considered; (ii) \textit{\acp{UAV} not moving towards the previous target position $\mathbf{d}_i[k-1]$}:~\ac{UAV} in moving towards the previous target are excluded; (iii) \textit{\acp{UAV} with sufficient path history $\mathbf{H}_{ij}[k]$}\footnote{It is worth noting that the proposed approach relies on a sequence of \ac{UAV} position estimates stored in the matrix $\mathbf{H}_{ij}[k]$, referred to as the path history matrix. Further details about the structure and algorithms for the update can be found in~\cite{Afzal2022Biob}}: Targets must have a path of at least three elements for path persistence analysis. 

In the second phase, if the~\ac{UAV} does not belong to the informed subgroup, it selects a target $\mathbf{d}_i[k]$ based on path similarity ($\sigma_{ijl}$) and persistence ($\gamma_{ij})$ metrics. Specifically, it chooses a target $j^\star$ from $\mathcal{T}_i[k]$ by maximizing a combined metric of path persistence and similarity. Thus, we can express this as:
\begin{equation}\label{eq:heuristicPresented}
    \mathbf{d}_i[k] = \Big[ \mathbf{H}_{ij^\star}[k] \Big]_1, 
\end{equation}
with
\begin{equation} \label{eq:t_select2}
    j^\star = \argmax_{j \in \mathcal{T}_i[k]} \left( \gamma_{ij} + \sum_{l \in \mset{T}_i[k] \backslash j} \sigma_{ijl} \right).
\end{equation}

Once the target is selected, the \ac{UAV} computes a path to it while avoiding obstacles and other \acp{UAV}. This is achieved by controlling the \ac{UAV} velocity, denoted as $\mathbf{u}_i$, which is the sum of navigation and collision avoidance vectors:
\begin{equation}\label{eq:controlSignal}
\mathbf{u}_i[k] = \mathbf{n}_i[k] + \mathbf{c}_i[k],
\end{equation}
where the navigation vector $\mathbf{n}[k]$ is determined by the \ac{UAV}'s informed status and proximity to neighbors. It guides the \ac{UAV} towards the desired target while maintaining cohesion with the swarm. On the other hand, the collision avoidance vector $\mathbf{c}[k]$ is a combination of vectors aimed at avoiding obstacles. It responds more assertively to nearby obstacles and is designed to ensure smooth motion.

In the rest of this section, we explain the process from the~\ac{UAV}'s perspective, and to simplify the notation, we drop the subscript $i$ from $\mathbf{n}_i[k]$ and $\mathbf{c}_i[k]$ variables. The navigation vector $\mathbf{n}[k]$ is given by:
\begin{equation}
\label{eq:n_vector}
    \mathbf{n}[k] = f_{ig}\mathbf{n}_I  + \bar{f}_{ig}\mathbf{n}_U,
\end{equation}
where $f_{ig} = 1$ if, and only if, the~\ac{UAV} has goal information, while $\bar{f}_{ig}$ is its complementary function. The vector $\mathbf{n}_I$ is used if the~\ac{UAV} is informed about the goal and is calculated as
\begin{equation}
\label{eq:nav_vec_informed}
    \mathbf{n}_I = \max \left( V^m, 1 - \frac{\sum\limits_{j \in \mathcal{N}_i} \lVert \mathbf{\check{p}}_{ij} - \mathbf{p}_i \rVert }{2R^f \lvert \mset{N}_i \rvert}  \right) K^n(\mathbf{a}_n - \mathbf{p}_i), 
\end{equation}
where, $\mset{N}_i$ is the set of neighboring~\acp{UAV}, $V^m \in (0,1)$ is the minimum normalized velocity of the informed~\ac{UAV}, and $K^n \in \mathbb{R}$ is a scaling coefficient to rescale the position vector to form the velocity control input ($\mathbf{u}_i[k]$). The second term in the $\max$ function depends on the average distance from the~\acp{UAV} in $\mathcal{N}_i$. The magnitude of vector $\mathbf{n}_I$ decreases as this average distance increases, preventing the informed~\ac{UAV} from wandering far away from the swarm. The $\max$ function ensures that the~\ac{UAV} always moves with a minimum velocity of $V^m$.

The collision avoidance vector for an obstacle $\mathbf{o}_r \in \mathcal{O}_i[k]$ is obtained by:
\begin{equation}\label{eq:collision_avoidance_vec}
    \mathbf{c}_r = \max \left( 0, \frac{1}{ \lVert \mathbf{p}_i - \mathbf{o}_r \rVert}  - \frac{1}{R^o} \right) \mathbf{\hat{c}}_r,
\end{equation}
with
\begin{equation}
    \mathbf{\hat{c}}_r = \argmax_{\mathbf{\hat{b}}\in\{\mathbf{\hat{c}}^+, \mathbf{\hat{c}}^-\}} \left( \frac{\mathbf{\hat{b}} \cdot \mathbf{u}_i[k-1]}{ \lVert \mathbf{u}_i[k-1] \rVert}   \right), 
\end{equation}
where vectors $\mathbf{\hat{c}}^+$ and $\mathbf{\hat{c}}^-$ denote two possible directions of motion to avoid collision with the obstacle $\mathbf{o}_r$. To maintain smooth motion, the vector with the least angular distance to the previous control input $\mathbf{u}_i[k-1]$ is used for collision avoidance. The magnitude of $\mathbf{c}_r$ is inversely proportional to the relative distance to the obstacle, meaning that the~\ac{UAV} reacts more strongly to nearby obstacles compared to farther ones. The collision control vector $\mathbf{c}[k]$ is a superposition of collision avoidance vectors of all the obstacles in $\mathcal{O}_i[k]$, given by:
\begin{equation}\label{eq:collision_avoidance}
    \mathbf{c}[k] = K^c  \sum_{\mathcal{O}_i[k]} \mathbf{c}_r,
\end{equation}
where $K^c$ is a scaling coefficient to rescale the summation.



\section{Experimental Results}
\label{sec:exp_results}

We evaluated the performance of the proposed approach through simulated experiments using Gazebo and field experiments conducted
in a natural forest. Videos showcasing these experiments are accessible at \url{https://mrs.felk.cvut.cz/pacnav}. Refer to \cite{MRS2022ICUAS_HW, Baca2021mrs} for additional information regarding the hardware and software used for the experiments. 


\begin{acronym}
    \acro{CNN}[CNN]{Convolutional Neural Network}
    \acro{DBSCAN}[DBSCAN]{Density-based Spatial Clustering of Applications with Noise}
    \acro{EKF}[EKF]{Extended Kalman Filter}
    \acro{FIFO}[FIFO]{First-In-First-Out}
    \acro{FOV}[FoV]{Field of View}
    \acro{GNSS}[GNSS]{Global Navigation Satellite System}
    \acro{ICNIRP}[ICNIRP]{International Commission on Non-Ionizing Radiation Protection}
    \acro{IMU}[IMU]{Inertial Measurement Unit}
    \acro{IR}[IR]{InfraRed}
    \acro{LiDAR}[LiDAR]{Light Detection and Ranging}
    \acro{LoS}[LoS]{Line of Sight}
    \acro{MAV}[MAV]{Micro Aerial Vehicle}
    \acro{ML}[ML]{Machine Learning}
    \acro{MPC}[MPC]{Model Predictive Control}
    \acro{MOCAP}[MoCap]{Motion Capture}
    \acro{PACNav}[PACNav]{Persistence Administered Collective Navigation}
    \acro{RMSE}[RMSE]{Root Mean Square Error}
    \acro{ROS}[ROS]{Robot Operating System}
    \acro{RTK}[RTK]{Real-Time Kinematic}
    \acro{SLAM}[SLAM]{Simultaneous Localization and Mapping}
    \acro{UAV}[UAV]{Unmanned Aerial Vehicle}
    \acro{UV}[UV]{UltraViolet}
    \acro{UVDAR}[UVDAR]{UltraViolet Direction And Ranging}
    \acro{UT}[UT]{Unscented Transform}
    \acro{UWB}[UWB]{Ultra-Wideband ranging}
    \acro{wrt}[w.r.t.]{with respect to}
\end{acronym}

\bibliographystyle{IEEEtran}
\bibliography{main}

\begin{thebibliography}{1}
\providecommand{\url}[1]{#1}
\csname url@samestyle\endcsname
\providecommand{\newblock}{\relax}
\providecommand{\bibinfo}[2]{#2}
\providecommand{\BIBentrySTDinterwordspacing}{\spaceskip=0pt\relax}
\providecommand{\BIBentryALTinterwordstretchfactor}{4}
\providecommand{\BIBentryALTinterwordspacing}{\spaceskip=\fontdimen2\font plus
\BIBentryALTinterwordstretchfactor\fontdimen3\font minus
  \fontdimen4\font\relax}
\providecommand{\BIBforeignlanguage}[2]{{%
\expandafter\ifx\csname l@#1\endcsname\relax
\typeout{** WARNING: IEEEtran.bst: No hyphenation pattern has been}%
\typeout{** loaded for the language `#1'. Using the pattern for}%
\typeout{** the default language instead.}%
\else
\language=\csname l@#1\endcsname
\fi
#2}}
\providecommand{\BIBdecl}{\relax}
\BIBdecl

\bibitem{Afzal2022Biob}
A.~{Ahmad} \emph{et~al.}, ``{PACNav: A collective navigation approach for UAV
  swarms deprived of communication and external localization},''
  \emph{Bioinspiration \& Biomimetics}, vol.~17, no.~6, pp. 1--19, 2022.

\bibitem{McGuireScienceRobotics2019}
K.~N. {McGuire} \emph{et~al.}, ``{Minimal navigation solution for a swarm of
  tiny flying robots to explore an unknown environment},'' \emph{Science
  Robotics}, vol.~4, no.~35, p. eaaw9710, 2019.

\bibitem{vicsek_2010_nature}
M.~Nagy \emph{et~al.}, ``{Hierarchical group dynamics in pigeon flocks},''
  \emph{Nature}, vol. 464, no. 7290, pp. 890--893, 2010.

\bibitem{olfati_2006_tac}
R.~{Olfati-Saber}, ``{Flocking for multi-agent dynamic systems: algorithms and
  theory},'' \emph{IEEE Transactions on Automatic Control}, vol.~51, no.~3, pp.
  401--420, 2006.

\bibitem{MRS2022ICUAS_HW}
D.~{Hert} \emph{et~al.}, ``{MRS Modular UAV Hardware Platforms for Supporting
  Research in Real-World Outdoor and Indoor Environments},'' in
  \emph{International Conference on Unmanned Aircraft Systems}, 2022, pp.
  1303--1312.

\bibitem{Baca2021mrs}
T.~Baca \emph{et~al.}, ``{The MRS UAV System: Pushing the Frontiers of
  Reproducible Research, Real-world Deployment, and Education with Autonomous
  Unmanned Aerial Vehicles},'' \emph{{Journal of Intelligent {\&} Robotic
  Systems}}, vol. 102, no.~26, pp. 1--28, 2021.

\end{thebibliography}

\end{document}